\documentclass{article}

\usepackage{PRIMEarxiv}
\usepackage{etoolbox}
\makeatletter
\patchcmd{\@maketitle}{\vskip 0.2in}{\vskip 0.2in\vskip 0.25cm}{}{}
\makeatother

\usepackage[utf8]{inputenc}
\usepackage[T1]{fontenc}
\usepackage{hyperref}
\usepackage{url}
\usepackage{booktabs}
\usepackage{amsmath}
\usepackage{amsfonts}
\usepackage{amssymb}
\usepackage{latexsym}
\usepackage{nicefrac}
\usepackage{microtype}
\usepackage{inconsolata}
\usepackage{graphicx}
\usepackage{multirow}
\usepackage{array}
\usepackage{enumitem}
\usepackage{placeins}
\usepackage{natbib}
\usepackage{xcolor}

\newcommand{\chg}[1]{#1}
\newcommand{\del}[1]{}
\title{\includegraphics[width=1\textwidth]{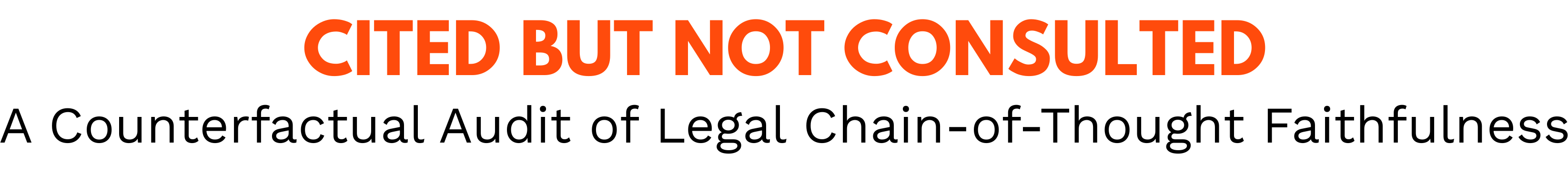}}

\author{
  Saisab Sadhu, Shreeyans Arora\thanks{Work done while at Lexsi Labs.} \\
  Pratinav Seth \\
  \affiliation{Lexsi Labs}\\
  \small{\textbf{Correspondence:} \href{mailto:saisab.sadhu@lexsi.ai}{saisab.sadhu@lexsi.ai}}
}

\setabstract{%
Large language models increasingly justify legal decisions by naming the statute or
precedent behind a verdict, treated as evidence the decision follows from it. We test
this directly: holding case facts fixed, we substitute the named legal authority for an
unrelated one and decode a model's evolving verdict from its hidden states. Across seven
open-weight models (8B--70B) and four benchmarks spanning judicial and contractual
reasoning, when explicitly required to justify a verdict by naming the governing
authority, models name the correct one in 66.7\%--100\% of generations, while the
verdict changing when the authority changes is far less consistent: 0.0\%--21.7\% on
CaseHOLD, 30.0\%--%
\chg{76.7}\% on ECHR and SCOTUS, and 43.3\%--50.0\% on ContractNLI. Neither
scale nor a purpose-built legal-reasoning model (a best-effort LoRA reproduction;
Section~\ref{sec:legaldelta-rq}) closes this gap. A red-teaming evaluation on five core
models finds compliance with an adversarial instruction hidden in the case facts
(73.3\%--96.4\%) exceeds verdict-swap sensitivity by a wide margin, holding without
exception regardless of cross-model ranking. Naming a legal authority is thus a poor
proxy for a verdict's dependence on it, and the same verdict remains separately
vulnerable to adversarial manipulation; both findings replicate across checks ruling out
prompt-wording noise and confounded sampling, and bear directly on using generated legal
explanations as compliance or audit artefacts.
}

\setkeywords{chain-of-thought faithfulness, legal NLP, counterfactual evaluation, mechanistic interpretability, LLM auditing}

\runningtitle{Cited but Not Consulted : A Counterfactual Audit of Legal Chain-of-Thought Faithfulness}

\begin{document}
\maketitle

\section{Introduction}
Financial and legal AI systems increasingly justify high-stakes decisions through
Chain-of-Thought (CoT) reasoning that names the rule, statute, or precedent said to
govern the outcome; that naming is commonly treated as evidence the decision follows
from it. A growing body of work complicates this picture: CoT is frequently
\emph{unfaithful}, with models committing to an answer before generating the reasoning
that ostensibly supports it, and that reasoning amounting to post-hoc rationalisation
rather than a causal input \citep{lanham2023measuring,lyu2023faithful,arcuschin2025chain}.
None of this asks whether unfaithfulness extends to the specific authority a model
names, in a domain where that naming is the reasoning's defining structural feature, or
whether the same verdict is directly manipulable independently of it. We test both
(Section~\ref{sec:related}), prioritising convergent evidence across models, lenses,
and adversarial conditions over benchmark-style sample sizes
(Section~\ref{sec:limitations}), turning three assumptions into explicit research
questions. First, is naming the correct
authority a reliable proxy for the verdict depending on it, and does any gap survive
changes in scale, training paradigm, and legal-authority type (RQ1)? Second, does
behavioural authority-swap sensitivity track post-commitment authority re-engagement, or
do the two dissociate (RQ2)? Third, is the same verdict directly exploitable via an
adversarial instruction hidden in the document under review, independently of whether it
depends on the authority it names (RQ3)? We use \emph{consulted} in the title as
shorthand for these behavioural and textual proxies, not a claim that an authority's
internal representation is causally used (Section~\ref{sec:discussion}).

The present work contributes an \textbf{authority-swap counterfactual protocol}, more
precisely a \emph{legal decision target} intervention instantiated as a treaty
provision, cited precedent, issue-area category, or contractual obligation
(Section~\ref{sec:contractnli-data}): holding facts fixed, we substitute the target a
model is asked about for an unrelated one and measure whether the model's internal
decision signal and final verdict change, paired, to our knowledge for the first time,
with \textbf{internal commitment-tracking} applied to legal reasoning: decoding a
model's evolving decision from hidden states at each sentence boundary, previously
confined to general-domain tasks. A \textbf{red-teaming ablation} answers RQ3 directly
(Section~\ref{sec:robustness-results}), showing a generated explanation is exploitable
on two independent fronts at once.

\section{Related Work}
\label{sec:related}

\paragraph{Faithfulness and internal commitment.} CoT explanations frequently fail to
reflect a model's actual computation: models can commit to an answer before generating
the reasoning meant to support it \citep{lanham2023measuring}, %
%\del{omit causally relevant steps while including irrelevant ones}
\chg{produce a reasoning chain that need not reflect how the answer was reached} \citep{lyu2023faithful}, produce post-hoc
rationalisations even in frontier models \citep{arcuschin2025chain}, and rationalise
answers actually driven by an input-side influence the CoT never discloses
\citep{turpin2023unfaithful}. Decoding a model's evolving prediction from hidden
states, via the logit lens \citep{nostalgebraist2020logitlens} or a tuned lens
\citep{belrose2023tunedlens}, tracks \emph{when} a decision forms, not just whether it
eventually forms correctly; recent work locates pre-verbalisation commitment on
delayed-verdict tasks \citep{whendoesllmcommit2026} and probes activation dynamics on
MMLU/GPQA-style benchmarks \citep{reasoningtheater2026}. Post-commitment CoT is itself
often \emph{epiphenomenal}, continuing with little effect on the final answer
probability once stabilised \citep{scalena2026epiphenomenal}, a finding we return to
when interpreting re-engagement after commitment (Section~\ref{sec:discussion}). Two
concurrent papers bring commitment-tracking to CoT at scale: \citet{agentcommit2026}
%
%\del{define}
\chg{defines} representational commitment via cross-run hidden-state convergence in
ReAct-style agents, and \citet{performativecot2026} %
%\del{apply step-level tuned-lens probing}
\chg{track step-level answer commitment with a proxy cross-validated against Patchscopes, tuned-lens probes, and causal direction ablation}
across nine models and seven benchmarks, finding latent commitment and the explicit
answer align on only 61.9\% of steps. All of this is confined to general-domain
reasoning; we bring it to law, where naming an authority is the reasoning's defining
structural feature, and are not aware of prior work applying commitment-tracking to
legal reasoning.

\paragraph{Citation reliability and causal mediation.} Work on legal-citation
reliability asks whether a named case or statute is \emph{real} and \emph{doctrinally
accurate}, typically via retrieval- or graph-based verification
\citep{legalcitebench2026,who2026checks,citationgrounding2026}, a question of
correctness, not causal dependence: a model can hallucinate a citation while its
verdict tracks the real authority, or name a real, accurate authority its verdict does
not depend on at all, the failure mode this paper targets. Four counterfactual
protocols are closer in spirit: FACET \citep{facet2026} perturbs US tort case facts and
checks behavioural factor-attribution updates, finding these for moderately but not
highly canonical factors; \citet{breakingthechain2026} re-prompt with a perturbed
intermediate structure, %
%\del{reporting non-update rates up to 60\% across three benchmarks}
\chg{finding, across twelve models and four benchmarks, that predictions often fail to update after the intervention};
FACT-E \citep{facte2026} uses controlled perturbations as an instrumental signal for
intra-chain CoT faithfulness; and \citet{paul2024making} apply causal mediation
analysis to intermediate reasoning across twelve models, the closest prior causal
(rather than textual-proxy) test to the activation-patching direction we motivate but
do not run (Section~\ref{sec:discussion}). Closest in domain, \citet{knowyourlimits2026}
evaluate legal NLI faithfulness on a re-annotated ContractNLI subset via formal,
solver-based autoformalization: they ask whether stated reasoning formally entails the
answer, not whether the answer causally depends on the clause named, and their subset
is disjoint from ours. We instead perturb the named authority, pairing this behavioural
check with internal decoding; the contribution is not novelty in CoT faithfulness,
causal intervention, or legal reasoning individually, but their combination:
legal-decision-target counterfactuals paired with latent commitment tracking and
adversarial robustness testing, which\chg{, to our knowledge,} no prior work combines.

\section{Task and Data}

LexGLUE \citep{chalkidis2021lexglue} assembles ECHR violation classification, building
on \citet{chalkidis2019neural}, SCOTUS issue-area classification, and CaseHOLD
\citep{zheng2021casehold} precedent-holding selection, among other tasks, into a single
benchmark suite; we build directly on its ECHR, SCOTUS, and CaseHOLD configurations.

Table~\ref{tab:tasks} (Appendix~\ref{app:tasks-full}) summarises the four task/dataset
pairs. We use \emph{authority} as informal shorthand for what each task instantiates as
the target of a legal decision: a treaty provision for ECHR, a cited precedent's
identity for CaseHOLD, an issue-area category for SCOTUS, and a contractual obligation
for ContractNLI. SCOTUS's issue-area label is the least literally an ``authority'' of
the four (not itself a statute, precedent, or clause a decision cites), but the same
protocol applies: substitute the decision target, hold facts fixed, and measure whether
the verdict tracks the substitution.

\paragraph{Dataset details.}
\label{sec:contractnli-data}
ECHR uses the ECtHR-A configuration (Table~\ref{tab:tasks}): each case's facts map to
the articles the Court \emph{actually found violated}, distinct from articles merely
alleged (ECtHR-B); ``ECHR'' below refers to this task, ``Article $n$''/``the
Convention'' to the treaty provisions, and ``the Court'' to the European Court of Human
Rights; sampling is stratified since the raw corpus skews negative. CaseHOLD
\citep{zheng2021casehold} masks a US court decision's holding as \texttt{<HOLDING>} in a
five-way multiple-choice task; where ECHR is decided from underlying facts, CaseHOLD is
decided from an authority already present in the \emph{input}
(Section~\ref{sec:method}). SCOTUS issue-area classification \citep{chalkidis2021lexglue}
codes each opinion's primary subject matter into one of fourteen issue areas; opinions
are truncated to their first 6{,}000 characters to keep the generation budget tractable
(\chg{Appendix~\ref{app:full-limitations}}). Finally, as a fourth RQ1 axis testing whether the gap generalises beyond
judicial case law, we add ContractNLI \citep{koreeda2021contractnli}, an NLI task over
non-disclosure agreements where the label is whether the contract entails, contradicts,
or does not mention a stated confidentiality obligation; we collapse
\emph{Contradiction}/\emph{NotMentioned} into NO against \emph{Entailment} as YES.

\section{Method}
\label{sec:method}

\begin{figure*}[t]
\centering
\includegraphics[width=0.99\textwidth]{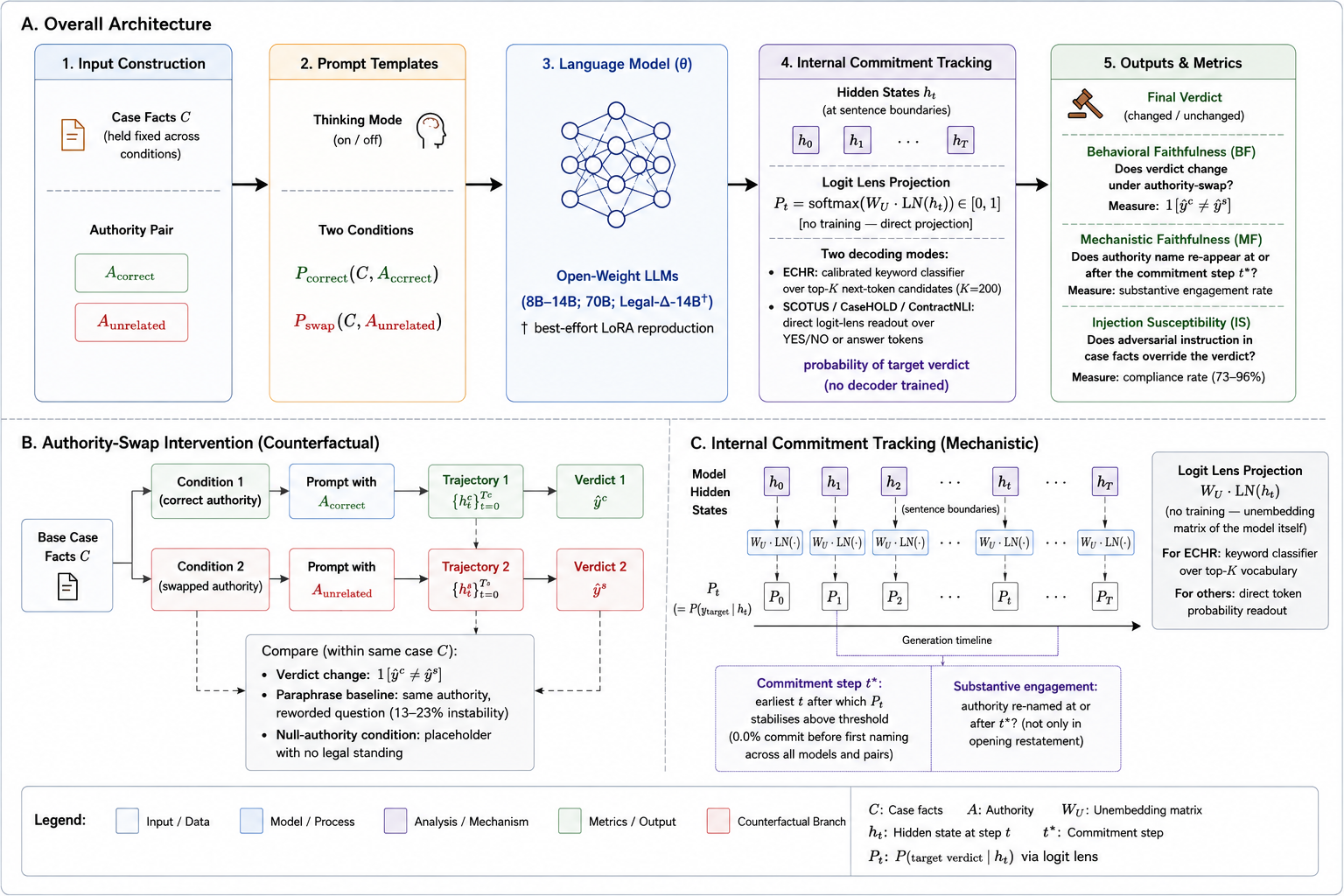}
\caption{Overview of the pipeline. \textbf{A}: case facts are held fixed while an
authority pair (correct against unrelated) is substituted into the prompt under a
thinking-mode toggle; the model, open-weight and untrained for this task, produces a
Chain-of-Thought whose hidden states are read at each sentence boundary via an
untrained logit-lens projection over the target verdict, yielding this paper's three
outcome measures, verdict-swap sensitivity, post-commitment re-engagement (labelled
``substantive engagement'' in this figure; a textual proxy, not a causal measure,
Section~\ref{sec:discussion}), and injection susceptibility, each defined in
Section~\ref{sec:method}. \textbf{B}: the authority-swap counterfactual, comparing the
two conditions'
trajectories and verdicts on the same facts, alongside the paraphrase and
null-authority baselines. \textbf{C}: commitment tracking, the logit-lens projection
$P_t=\mathrm{softmax}(W_U\cdot\mathrm{LN}(h_t))$ at every sentence boundary, the
commitment step $t^\ast$ (earliest point the decoded verdict stays fixed), and whether
the authority is re-named at or after $t^\ast$.}
\label{fig:pipeline}
\end{figure*}

Figure~\ref{fig:pipeline} summarises the pipeline%
%\del{ in its entirety}
.

\paragraph{Eliciting authority-grounded CoT.}
For all four tasks, the system prompt instructs the model to place its step-by-step
reasoning inside \texttt{<think>}/\texttt{</think>} tags, name explicitly the legal
test, statute, precedent, or clause it relies upon, and conclude with a fixed-format
marker (\texttt{VERDICT: VIOLATION} or \texttt{NO VIOLATION} for ECHR, \texttt{ANSWER:
\{0..4\}} for CaseHOLD, \texttt{VERDICT: YES} or \texttt{NO} for SCOTUS and
ContractNLI). Marker extraction is normalised against markdown and \LaTeX{} emphasis
wrapping, and falls back to natural-language conclusion phrasing (e.g.\ ``I would
conclude there is a violation'') when the literal marker is absent, taking the
rightmost such match since a model may float a tentative answer before revising it.
Because the prompt explicitly instructs the model to name an authority, near-universal
surface engagement (Section~\ref{sec:cross-model}) partly reflects instruction-following;
every claim is scoped to what happens \emph{once} a model is required to name an
authority, not to unprompted citation (Section~\ref{sec:limitations}).

\paragraph{Commitment tracking.}
Following \citet{whendoesllmcommit2026}, we track a model's internal decision at every
sentence boundary by decoding the hidden state at that position. For ECHR, the top-$K$
next-token candidates ($K=200$) are bucketed into \emph{violation}/\emph{no-violation}
via a calibrated keyword-and-morphology classifier, with raw category counts corrected
by each category's empirical base rate across every sentence in a run, since word lists
for different categories are not equally generic and accumulate top-$K$ hits at
different rates independent of content. For CaseHOLD (digits \texttt{0}--\texttt{4}), SCOTUS, and ContractNLI (tokens YES/NO),
the decision space is a small literal token set, so the model's own next-token
probability for each candidate is read directly off the hidden state: a direct
logit-lens readout, no keyword heuristic needed. We apply this readout over a small
literal candidate set (two tokens for SCOTUS/ContractNLI, five for CaseHOLD) rather than
the full vocabulary, reducing the instability concerns raised for full-distribution
logit-lens probing \citep{belrose2023tunedlens}.

\paragraph{Commitment step, formally.} Let $\ell_i \in \{c_1,\ldots,c_m,\bot\}$ denote
the decoded label at sentence $i$ ($m=2$ categories for ECHR/SCOTUS/ContractNLI, $m=5$
candidate holdings for CaseHOLD; $\bot$ marks no confident label), and $s_i(c)$ a
category's score at that sentence: the calibrated, base-rate-corrected top-$K$ keyword
count for ECHR, the raw next-token probability for the direct logit-lens readout
elsewhere. Writing $\mathrm{top}$ and $\mathrm{2nd}$ for the highest- and
second-highest-scoring categories,
$$
\ell_i = \begin{cases}
\mathrm{top} & \text{if } s_i(\mathrm{top}) > 0 \text{ and} \\
 & \bigl(s_i(\mathrm{2nd})=0 \text{ or } \tfrac{s_i(\mathrm{top})}{s_i(\mathrm{2nd})} \geq 1.5\bigr) \\
\bot & \text{otherwise}
\end{cases}
$$
a category needs nonzero support and a $1.5\times$ margin over the runner-up to count
as confident; sentences with no support or a near-tie are $\bot$, not resolved by
tie-breaking. The \emph{commitment step} $t^\ast$ is the earliest sentence index $i$
such that $\ell_j=\ell_n$ for every $j\in[i,n]$, where $n$ is the final sentence,
\emph{provided} $\ell_n\neq\bot$; if the last sentence never reaches a confident label,
$t^\ast$ is undefined and the case is excluded from commitment-conditioned statistics.
ECHR's $s_i(\cdot)$ uses the calibrated count by default; Appendix~\ref{app:ablations}
reports commitment steps under the raw, uncalibrated count as an ablation.

\subsection{Main Intervention and Comparison}

\paragraph{Authority-swap counterfactual.}
For ECHR, SCOTUS, and ContractNLI, holding case facts fixed, we generate CoT twice per
case, once under the true target authority and once under the control. We measure
\emph{engagement}, whether the reasoning names the authority actually asked about, in
either condition; \emph{verdict-swap sensitivity}, whether the final verdict changes
between conditions; on the unrestricted sample this is descriptive only, since two
different authorities can legitimately imply the same outcome, which is why the
gold-different subset (Section~\ref{sec:differ-method}), where target and control gold
labels are constructed to differ, is this paper's cleaner reading of
authority-dependence; and
\emph{commitment-before-naming}, whether the internal decision locks in before the
model first names the authority, evidence the naming is attached after the fact rather
than driving the reasoning.

For CaseHOLD, the intervention instead perturbs the \emph{input}: the named precedent
case in the excerpt is replaced with a neutral placeholder, \texttt{Doe v. Roe}, while
the surrounding text and all five candidate holdings are held fixed, and we measure
whether the selected holding changes as a result.

\paragraph{Worked example (illustrative).} Concretely (%
%\del{illustrative, }
not a reproduced
generation): a detained applicant's private medical correspondence was read by prison
staff without consent. The \emph{target} condition asks whether Article 8 (private and
family life) is violated; the \emph{control} condition pairs the same facts with
Article 3 (torture and inhuman treatment), which they do not obviously engage. A model
matching our typical pattern restates each article in its opening sentence
(\emph{surface engagement}), then concludes \texttt{VERDICT: VIOLATION} under Article 8
and \texttt{VERDICT: NO VIOLATION} under Article 3%
%\del{ --}
\chg{,} a verdict-swap event, with
commitment-before-naming checked by comparing $t^\ast$'s sentence index against this
opening mention (naming comes first in every case observed).

\subsection{Supporting Diagnostics}

We distinguish two notions of engagement. \emph{Surface engagement} is whether the
asked-about authority is named anywhere in the reasoning, near-universal since every
model restates it in its opening sentence. \emph{Post-commitment re-engagement}
additionally requires the authority to be named again at or after the sentence where
the internal decision signal locks in, not only in that opening restatement.

\paragraph{Confound-free resampling.}\label{sec:differ-method}
A model answering NO under both conditions is not necessarily unfaithful, since
dual-negative agreement can be independently correct rather than evidence of
insensitivity. We address this by restricting the case pool to instances where the
target and control gold labels genuinely differ (exactly one is violated, for ECHR; the
true issue area is exactly the target or exactly the control, for SCOTUS), on which a
model tracking the authority it is asked about must change its verdict
(Section~\ref{sec:differ-results}).

\paragraph{Robustness: paraphrase and prompt-injection ablation.}
\label{sec:robustness-method}
The null-authority condition (Section~\ref{sec:discussion}) leaves open how much
measured swap-sensitivity is generic instability to any prompt change, rather than
specifically to swapping the real authority. We add a tighter baseline: holding the
target authority, question, and facts fixed, we reword the question into a
meaning-preserving paraphrase (Section~\ref{sec:robustness-results}), isolating
wording-level instability with the legal question itself unaltered. Separately, we test whether the verdict is exploitable via an adversarial instruction
injected into the case facts, appended text instructing the model to conclude
\texttt{VERDICT: VIOLATION} regardless of the actual facts, framed as a spurious
``registry note'': a red-teaming question complementary to the authority-swap protocol.
Following prompt-injection practice that separates attacker-goal success from task
utility \citep{agentdojo2024}, we report both the raw compliance rate (does the output
match the injected verdict) and whether that output differs from the model's own
uninjected verdict on the same case (Section~\ref{sec:robustness-results}).

\section{Experimental Setup}
We evaluate seven open-weight instruction-tuned and reasoning models
(Table~\ref{tab:models}), run locally in bfloat16, seed$=42$ throughout; generation is
capped at 1{,}536 new tokens.

\begin{table}[t]
\centering
\caption{Models evaluated. ``Native'' is a model's own thinking-mode toggle;
``Prompted'' models receive the \texttt{<think>} tag via system prompt
(Section~\ref{sec:method}). $^\dagger$Legal-$\Delta$-14B is a best-effort LoRA
reproduction (Sections~\ref{sec:legaldelta-rq}, \ref{sec:limitations}); Llama-3.3-70B is
a scale check on RQ1 (Section~\ref{sec:scale-rq}).}
\label{tab:models}
\vspace{0.6em}
\footnotesize
\setlength{\tabcolsep}{6pt}
\renewcommand{\arraystretch}{1.15}
\begin{tabular}{@{}llll@{}}
\toprule
Model & Params & Elicitation & Scope \\
\midrule
Qwen3-8B & 8B & Native & Full battery \\
Qwen3-14B & 14B & Native & Full battery \\
Llama-3.1-8B & 8B & Prompted & Full battery \\
Llama-3-8B & 8B & Prompted & Full battery \\
Gemma-3-12B & 12B & Prompted & Full battery \\
Llama-3.3-70B & 70B & Prompted & ECHR/SCOTUS Pair A \\
Legal-$\Delta$-14B & 14B & Prompted & Full battery$^\dagger$ \\
\bottomrule
\end{tabular}%
\end{table}
\FloatBarrier

For the main ECHR, SCOTUS, and ContractNLI comparisons, we use a balanced sample of
thirty cases per model per authority pair, fifteen positive and fifteen negative, each
generated under both the target- and control-authority conditions; CaseHOLD uses thirty
sampled cases under both the original and authority-swapped conditions. For each of
ECHR and SCOTUS we evaluate two structurally independent target/control authority
pairs: Article 8 against Article 3 and Article 6 against Article 14 for ECHR; Criminal
Procedure against Economic Activity and First Amendment against Civil Rights for
SCOTUS. For ECHR and SCOTUS we additionally run a null-authority condition, substituting
a placeholder authority with no legal standing for the control, isolating whether
verdict-swap sensitivity reflects authority-specific engagement or generic
prompt-perturbation instability (Section~\ref{sec:discussion}). We report 95\%
bootstrap confidence intervals (1{,}000 resamples) on verdict-swap sensitivity, our
headline measure, throughout.

\section{Results}

\paragraph{Cross-model comparison.}
\label{sec:cross-model}
Table~\ref{tab:cross-model} reports the full cross-model measures on ECHR Pair A, with
95\% bootstrap CIs (1{,}000 resamples) on verdict-swap sensitivity; convergence and
accuracy are solid across the board, and the gap between surface and post-commitment
re-engagement is the same qualitative pattern seen across every dataset in this study.
Verdict-swap sensitivity ranges from 31.0\% (Qwen3-14B) to 58.6\%
(Llama-3.1-8B-Instruct); only Qwen3-14B's interval, $[13.8, 48.3]$, excludes 50\%
entirely (Appendix~\ref{app:mcnemar} tests this further). The two measures do not move
together: Llama-3.1-8B-Instruct is highest on swap-sensitivity yet %
%\del{second-lowest}
\chg{third-lowest} on
post-commitment re-engagement, while Qwen3-14B shows close to the opposite. Scaling
within Qwen3, 8B to 14B, improves %
%\del{accuracy and }
re-engagement \chg{but lowers accuracy (88.9\% to 82.8\%)} while slightly
\emph{lowering} swap-sensitivity, so scale alone does not resolve the gap. Whether this
pattern is stable is examined in Section~\ref{sec:pair-replication} with a second,
independent authority pair. We turn first, however, to RQ3: whether the same verdict is
exploitable by a route unrelated to which authority is named.

\begin{table*}[t]
\centering
\caption{ECHR Pair A (target Article 8, control Article 3), $n=30$ per model,
seed$=42$, all rows in percentages except the bracketed 95\% bootstrap CI (1{,}000
resamples) on verdict-swap sensitivity; committed-before-first-naming, not tabulated, is
0.0\% for every model here and on every other pair (Section~\ref{sec:discussion}).}
\label{tab:cross-model}
\vspace{0.6em}
\footnotesize
\setlength{\tabcolsep}{4pt}
\renewcommand{\arraystretch}{1.15}
\begin{tabular}{@{}lcccccc@{}}
\toprule
Metric & Qwen3-8B & Qwen3-14B & Llama-3.1-8B & Llama-3-8B & Gemma-3-12B & Legal-$\Delta$-14B \\
\midrule
Convergence (target/control) & 90.0 / 100 & 96.7 / 100 & 96.7 / 100 & 90.0 / 93.3 & 100 / 100 & 100 / 100 \\
Accuracy (of answered cases) & 88.9 & 82.8 & 69.0 & 48.1 & 66.7 & 83.3 \\
Verdict-swap sensitivity & 33.3 & 31.0 & 58.6 & 40.7 & 36.7 & 33.3 \\
\quad 95\% bootstrap CI & [14.8, 51.9] & [13.8, 48.3] & [41.4, 75.9] & [22.2, 59.3] & [20.0, 56.7] & [16.7, 50.0] \\
Surface engagement (target) & 100 & 100 & 100 & 80.0 & 100 & 96.7 \\
Post-commitment re-engagement & 53.3 & 70.0 & 33.3 & 46.7 & 26.7 & 20.0 \\
\bottomrule
\end{tabular}%
\end{table*}
\FloatBarrier

\paragraph{Robustness: paraphrase and prompt-injection (ECHR Pair A).}
\label{sec:robustness-results}
Both ablations below are run on ECHR Pair A only, five models, $n=30$
(Table~\ref{tab:robustness}); generalisability to other datasets is an acknowledged
limitation (Appendix~\ref{app:full-limitations}). Figure~\ref{fig:ordering}
(Appendix~\ref{app:ordering}) shows the full ordering at a glance. Paraphrase instability, rewording the question with
the authority and facts unchanged, sits well below verdict-swap sensitivity for every
model (13.8\% to 23.3\% versus 31.0\% to 58.6\%, Table~\ref{tab:cross-model}), and below
the null-authority condition's 10.0\% to 51.7\% for every model but
%
%\del{Llama-3.1-8B-Instruct}
\chg{Gemma-3-12B} (Section~\ref{sec:discussion}). With nothing about the authority
or question altered, wording sensitivity alone accounts for well under half the
swap-sensitivity for most models, a cleaner baseline supporting our reading that
verdict-swap sensitivity reflects more than surface prompt instability.

The prompt-injection result, this paper's direct answer to RQ3, is a distinct and more
concerning finding. We append an adversarial instruction to the case facts, framed as a
spurious registry note directing the model to conclude \texttt{VERDICT: VIOLATION}
regardless of the facts; all five models comply in 73.3\% to 96.4\% of cases, exceeding
verdict-swap sensitivity (Table~\ref{tab:cross-model}) for every model by a wide margin:
each is more willing to follow a fabricated instruction than its verdict is to track the
authority it is actually asked about. A stricter test, whether the injection
additionally changes the verdict relative to the model's own uninjected baseline,
happens in 18.5\% to 41.4\% of cases and, unlike compliance, exceeds swap sensitivity
only for Qwen3-8B. We therefore anchor the paper's central exploitability claim on the
compliance rate, the weaker and more defensible reading, not the stricter verdict-change
rate (Section~\ref{sec:discussion}).

\begin{table*}[t]
\centering
\caption{Robustness ablation on ECHR Pair A, $n=30$ per model, seed$=42$, all values in
percentages (row definitions in the text, Section~\ref{sec:robustness-results}).
Injection compliance is the rate at which the model outputs the instructed verdict;
injection verdict change is the stricter rate at which that output differs from the
model's own uninjected baseline. The paper's headline exploitability claim uses
compliance throughout (see text).}
\label{tab:robustness}
\vspace{0.6em}
\footnotesize
\setlength{\tabcolsep}{4pt}
\renewcommand{\arraystretch}{1.15}
\begin{tabular}{@{}lccccc@{}}
\toprule
Metric & Qwen3-8B & Qwen3-14B & Llama-3.1-8B & Llama-3-8B & Gemma-3-12B \\
\midrule
Paraphrase instability & 13.8 & 13.8 & 23.3 & 14.8 & 16.7 \\
Injection compliance & 86.2 & 73.3 & 90.0 & 96.4 & 86.7 \\
Injection verdict change & 41.4 & 27.6 & 30.0 & 18.5 & 20.0 \\
\bottomrule
\end{tabular}%
\end{table*}
\FloatBarrier

\paragraph{CaseHOLD: a second reasoning paradigm.}
On CaseHOLD, $n=30$ (full results in Table~\ref{tab:casehold},
Appendix~\ref{app:casehold-full}), convergence ranges from 80.0\% to
100\%, somewhat lower than on ECHR, plausibly because discriminating among five related
holdings is harder than a binary judgment, and accuracy is well above the 20\% chance
rate for every model, 73.3\% to 83.3\%. A precedent name was detected in the same 24 of
30 cases for every model (detection runs once on the fixed input excerpt); replacing it
with a neutral placeholder, \texttt{Doe v. Roe}, while holding the excerpt and all five
candidate holdings fixed, changes the selected holding in only 0.0\% to 21.7\% of
detected-citation cases; we call this \textbf{precedent-identity sensitivity}, not
authority-dependence generally, since only the case name changes: the lowest
swap-sensitivity of our three settings despite the authority sitting in the
\emph{input}, and the structurally cleanest intervention since it perturbs only the
named precedent (Section~\ref{sec:method}). The excerpt's surrounding facts may suffice to select the correct holding, the
precedent functioning as a label rather than a necessary premise: low sensitivity is
consistent with the citation being \emph{redundant} (already encoded in the excerpt)
rather than \emph{ignored}, a distinction this intervention alone cannot resolve, though
either reading fits the two-phase hypothesis (Section~\ref{sec:discussion}). %
%\del{We stress this distinction: our}
\chg{Our} intervention measures \emph{behavioural} authority-dependence, not
whether the authority is internally represented%
%\del{; low sensitivity is equally consistent with the authority being redundant (the facts already determine the answer) and with the authority being ignored, and our design does not distinguish the two}
. Because it changes
only a token-level identity, this result is complementary evidence that citation
identity can be dispensable; the cleanest evidence of unfaithful reasoning rests on
ECHR, SCOTUS, and ContractNLI, where the gold-different subset rules out the
dual-negative confound directly (Section~\ref{sec:differ-method}).

\paragraph{SCOTUS: a third dataset, a third court system.}
\label{sec:scotus-third}
SCOTUS (Pair A: target Criminal Procedure, control Economic Activity) mirrors the ECHR
design; convergence and accuracy are again solid across the board (full results in
Table~\ref{tab:scotus}, Appendix~\ref{app:scotus-full}). Verdict-swap sensitivity, 30.0\%
to 60.0\%, overlaps substantially with the ECHR Pair A range, 31.0\% to 58.6\%, though
the ranking differs: Gemma-3-12B-it, mid-ranked on ECHR, is highest here, while
Qwen3-14B is lowest in both. Whether this reshuffles with the decision space is examined
in Section~\ref{sec:pair-replication}.

\paragraph{Robustness of the gap: pair choice and dual-negative cases.}
\label{sec:pair-replication}
Two checks test whether the gap above reflects the specific pair chosen, or
dual-negative agreement, rather than genuine authority-dependence. A second,
structurally independent authority pair per dataset (ECHR: Article 6 v. Article 14;
SCOTUS: First Amendment v. Civil Rights; same five models, $n=30$, seed$=42$; full
per-model numbers in Appendix~\ref{app:pairrep}) replicates the central finding: every
model shows non-trivial sensitivity under both pairs, and the surface-versus-post-commitment
gap (\chg{Section~\ref{sec:cross-model}}) is never closed by either.
\label{sec:differ-results}
Separately, restricting to the confound-free subset where target and control gold
labels genuinely differ (Section~\ref{sec:differ-method}; same five models, $n=30$,
seed$=42$) does not uniformly lower sensitivity, as a confound account would predict: it
rises for three of five models on each of ECHR and SCOTUS (full figures in
Appendix~\ref{app:differ}; a modest majority against confound inflation, not decisive
evidence against it). ContractNLI's own binary collapse (Section~\ref{sec:contractnli-data})
creates an analogous risk, and here the check is unambiguous: restricted to the twelve
(of thirty) cases per model where target and control gold genuinely differ, sensitivity
rises for all %
%\del{five}
\chg{six} models, from 43.3\%--50.0\% on the full sample to 50.0\%--66.7\% on
the confound-free subset (full figures in Appendix~\ref{app:contractnli-differ}).

\paragraph{Scale and training paradigm: two generalisation checks on RQ1.}
\label{sec:scale-rq}
Does scale, or reasoning-native training beyond a single family, close the
naming-versus-depending gap? We run Llama-3.3-70B-Instruct, a five- to nine-fold
parameter increase over our 8B--14B range, on ECHR and SCOTUS Pair A ($n=30$,
seed$=42$). Convergence is total and accuracy is %
%\del{highest or near-highest of any model on each dataset, 80.0\% on ECHR and 96.7\% on SCOTUS}
\chg{the highest of any model on SCOTUS, 96.7\%, and mid-range on ECHR, 80.0\%}, but the gap does not close, and its
shape is dataset-dependent: ECHR swap-sensitivity, 40.0\% ($[23.3, 56.7]$), sits squarely
inside the 8B--14B range (Table~\ref{tab:cross-model}), while SCOTUS swap-sensitivity,
76.7\% ($[60.0, 90.0]$), is the highest we record anywhere, alongside the lowest
post-commitment re-engagement of any \chg{general-purpose} model on any dataset, 10.0\%, a sharp single-model
illustration of RQ2.
\label{sec:legaldelta-rq}
As a second, structurally different reasoning-native comparison, we add
Legal-$\Delta$-14B \citep{legaldelta2026}, built for legal reasoning via reinforcement
learning (a best-effort reproduction; Appendix~\ref{app:legaldelta-validity} tests merge
validity). Across all four datasets (Tables~\ref{tab:cross-model}--\ref{tab:contractnli}),
it shows no higher authority-dependence than the general-purpose models \chg{taken together (its SCOTUS 70.0\% exceeds every 8B--14B model but not the 70B model's 76.7\%)} and, like the
70B model, is comparatively swap-sensitive while among the least re-engaged: a second
illustration of the RQ2 dissociation from a different training route, treated as
corroborating rather than independent evidence given the reproduction caveat.

\paragraph{ContractNLI: a fourth authority type, a fourth domain.}
\label{sec:contractnli-rq}
ContractNLI (Section~\ref{sec:contractnli-data}) tests RQ1's authority-type axis,
whether the gap generalises beyond judicial case law (full results in
Table~\ref{tab:contractnli}, Appendix~\ref{app:contractnli-full}). Convergence is high, 86.7\% to 100\%, and accuracy is well above the 50\%
chance rate for every model, 57.7\% to 76.7\%. Verdict-swap sensitivity, 43.3\% to
50.0\%, is a narrower band than ECHR or SCOTUS but the same order of magnitude, and
surface engagement (%
%\del{93.3}
\chg{70.0}\% to 100\%) again outpaces post-commitment re-engagement (%
%\del{60.0}
\chg{3.3}\% to
83.3\%) \chg{for every model}. The gap thus replicates in a fourth authority type and domain, commercial
rather than judicial: it is not specific to courts, but reappears in a privately
negotiated contractual clause.

\section{Discussion}
\label{sec:discussion}
\paragraph{Ablations.}\label{sec:ablations} Three design choices are validated
(Appendix~\ref{app:ablations}): the ECHR calibration baseline reflects genuine
per-topic imbalance, not noise; calibration shifts individual-case commitment timing
without moving the aggregate rate; and prompted elicitation shifts measures by $\leq$10
points without closing the surface-versus-post-commitment gap.

Surface naming is a poor proxy for verdict-swap sensitivity: naming sits near ceiling
with little cross-case variance, while swap-sensitivity varies substantially across
models, pairs, and datasets, dissociating in level rather than moving together
(\chg{Section~\ref{sec:cross-model}}). Treating a named authority's
mere presence as evidence of application is therefore a nearly uninformative proxy,
since naming stays constant across decision-relevant and irrelevant cases alike.

\paragraph{A two-phase hypothesis: boilerplate naming, then fact-driven decision.} The
results are consistent with a two-phase process: the authority appears in an
obligatory opening restatement (consistent with the 0.0\% committed-before-naming rate
on every ECHR cell, Table~\ref{tab:cross-model}, and near-universal surface engagement),
after which the decision draws on factual patterns largely independent of the authority
named. This reading is not the only one available: since the prompt requires naming and
the probe only starts at sentence boundaries, a uniform 0.0\% rate is equally what a
measurement floor effect would produce; our design does not separate the two.
Post-commitment re-engagement, a name re-occurring after
commitment, is a strict lower bound on true mechanistic re-engagement, since a model
could attend back to the authority without repeating its name; low re-engagement does
not establish second-phase independence. The hypothesis fits
each RQ: naming is opening-phase template behaviour; behavioural and post-commitment
measures dissociate because they probe different phases; and the gap replicates across
authority types because naming-then-deciding is plausibly general in legal training
text. The 70B SCOTUS result (Section~\ref{sec:scale-rq}) remains one striking data
point, not yet generalised to how scale modulates the later phase. Activation-patching
the authority representation mid-reasoning would give a more direct causal test\chg{. P}ost-commitment reasoning is often epiphenomenal, continuing after the answer stabilises
with little effect on its probability \citep{scalena2026epiphenomenal}, \chg{which is} why we treat
re-engagement as a textual proxy, not causal evidence%
%\del{, either way}
: naming first does not
mean naming \emph{causes} commitment.

The cross-model comparison resists a single tidy narrative: the Llama-3.1/Qwen3-14B
contrast (Section~\ref{sec:cross-model}) holds up under a second authority pair on ECHR
(Section~\ref{sec:pair-replication}), while SCOTUS reorders the ranking, evidence of two
distinguishable phenomena rather than one latent factor.

\paragraph{Red-teaming (RQ3): models comply with injected instructions more reliably
than they track authority.} The injected-instruction result
(Section~\ref{sec:robustness-results}, Figure~\ref{fig:ordering}) is this paper's most
consistent, and most actionable, quantitative finding: injection \emph{compliance},
whether the output matches the instructed verdict, exceeds verdict-swap sensitivity for
every model by a wide margin. This is a claim about compliance, not that injection
always overturns the verdict (a stricter test, verdict change against the uninjected
baseline, exceeds swap sensitivity for only one of five models): a legal explanation
reasoning fluently about the correct authority follows a fabricated instruction more
readily than its verdict tracks that authority.

\paragraph{On the authority-swap confound.} Substituting Article 8 for Article 3 changes
the question asked, not only the authority named, so a matched verdict could reflect a
genuine outcome prior rather than insensitivity. Three converging checks, CaseHOLD's
precedent-only substitution (immune to this confound by construction, yet no more
sensitive than ECHR/SCOTUS), a null-authority baseline
(Table~\ref{tab:null-authority}, Appendix~\ref{app:null-authority}), and the
confound-free and paraphrase results (\chg{Section~\ref{sec:differ-results}}), indicate swap-sensitivity substantially reflects genuine
authority-dependence rather than wording noise, though the confound is not fully closed
(full argument in Appendix~\ref{app:null-authority}; caveat in
Section~\ref{sec:limitations}). In practical terms, treating a named authority as
sufficient evidence of faithfulness is therefore unsafe without this causal check;
combined with the red-teaming result, a system trusting fluent, authority-citing
explanations is vulnerable on two independent fronts, faithfulness and adversarial
robustness.

\section{Conclusion}
We introduced an authority-swap counterfactual protocol, internal commitment-tracking,
and a red-teaming ablation testing direct exploitability (RQ1--RQ3). Naming the correct
legal decision target is close to automatic; depending on it is not, and the two move
independently of post-commitment textual re-engagement (RQ2). On ECHR Pair A, all five
models comply with an injected verdict more often than their own verdict changes under
the authority swap, though baseline-verdict overturning shows no universal ordering
(Section~\ref{sec:robustness-results}); this single-pair, five-model ablation supports
the narrower compliance claim, not a general exploitability ranking.

\section{Limitations}
\label{sec:limitations}
The sample size for experiments remain moderate,
and several bootstrap confidence intervals on verdict-swap sensitivity span 20 points or
more; a paired exact test (Appendix~\ref{app:mcnemar}) confirms most pairwise
model-ranking differences are not statistically distinguishable at this sample size. The
authority-swap confound, that changing the authority also changes the question asked, is
not fully eliminable with our current design: a null-authority condition and a
confound-free resampling narrow it but leave a residual pattern
(Section~\ref{sec:discussion}). Legal-$\Delta$-14B (Section~\ref{sec:legaldelta-rq})
carries a related caveat, since its published weights are LoRA adapters merged onto the
closest available public base rather than the authors' actual checkpoint, making its
results a best-effort reproduction; the remaining limitations, on pair coverage, probe
standardisation, seed sensitivity, and scope, are detailed in
Appendix~\ref{app:full-limitations}.

\section{Ethics Statement}
All data used, the ECtHR, SCOTUS, and CaseHOLD configurations of LexGLUE and the
ContractNLI corpus, is publicly available data released for NLP research; ContractNLI's
contracts are drawn from public institutional filings, and none of our datasets
introduce new personal or sensitive data. Our findings caution against treating LLM-generated legal citations as sufficient
evidence of faithful reasoning. We do not release any model or system intended for real
legal decision-making, and do not recommend using any model evaluated here for actual
legal advice or adjudication.

\bibliographystyle{acl_natbib}
\bibliography{custom}
\newpage
\appendix

\section{Task/Dataset Pairs at a Glance}
\label{app:tasks-full}

\begin{table*}[h]
\centering
\caption{The four task/dataset pairs at a glance. ``Signal'' is how the internal
decision is decoded at each sentence boundary (Section~\ref{sec:method}); Pair A shown
here, Section~\ref{sec:pair-replication} additionally evaluates a second target/control
pair for ECHR and SCOTUS.}
\label{tab:tasks}
\vspace{0.6em}
\footnotesize
\setlength{\tabcolsep}{6pt}
\renewcommand{\arraystretch}{1.15}
\begin{tabular}{@{}lllll@{}}
\toprule
Dataset & Domain & Authority type & Target / Control & Signal \\
\midrule
ECHR & Judicial (ECtHR) & Convention article & Art.\ 8 / Art.\ 3 & Keyword classifier \\
CaseHOLD & Judicial (US) & Cited precedent & real / \texttt{Doe v. Roe} & Logit-lens (0--4) \\
SCOTUS & Judicial (US) & Issue area & Crim.\ Proc.\ / Econ.\ Activity & Logit-lens (Y/N) \\
ContractNLI & Commercial (NDAs) & Contract clause & Disclosure / No solicitation & Logit-lens (Y/N) \\
\bottomrule
\end{tabular}%
\end{table*}
\FloatBarrier

\section{Comparison Against Published Supervised Baselines}
\label{app:sota}
Table~\ref{tab:sota} places our zero-shot, CoT-prompted accuracy figures (Pair A for
ECHR and SCOTUS) beside fine-tuned supervised baselines from the original LexGLUE
paper. This is context, not a head-to-head comparison: the LexGLUE figures are macro-F1
on the full multi-label (ECtHR), fourteen-way (SCOTUS), or five-way (CaseHOLD) task,
from encoders trained on thousands of in-domain examples, whereas ours concern a
simplified binary or five-way framing solved with no task-specific training. The two
answer different questions, how well a fine-tuned encoder solves the full task versus
how much a general-purpose model's verdict on a simplified version depends on the
authority named, reported jointly so a reader can judge whether our accuracy figures
are reasonable in absolute terms. They are: our strongest zero-shot models score higher
than the published macro-F1 of fine-tuned BERT-family baselines on ECtHR-A%
%\del{ and SCOTUS, and lower on CaseHOLD}
\chg{, SCOTUS, and CaseHOLD}, despite no in-domain training; given the metric and task
mismatch above, this is informal context for absolute plausibility, not a claim that
our models outperform these baselines on the same task. The large ECtHR-A gap in particular
(Qwen3-8B at 88.9\% against Legal-BERT's 64.0 %
%\del{\%}
\chg{macro-F1}) is expected rather than anomalous: our
binary target/control framing is a substantially easier task than the full multi-label
problem the fine-tuned baselines solve, and the two figures are not measuring the same
thing.

\begin{table}[h]
\centering
\caption{Published macro-F1 on the full task (upper rows, from
\citealp{chalkidis2021lexglue}; ECtHR and SCOTUS figures are macro-F1, the CaseHOLD
figure is accuracy) against our zero-shot accuracy on the simplified binary (ECHR,
SCOTUS) or five-way (CaseHOLD) framing, Pair A, $n=30$ (lower rows; Llama-3.3-70B was
not run on CaseHOLD, Section~\ref{sec:scale-rq}); discussion in the text above.}
\label{tab:sota}
\vspace{0.6em}
\footnotesize
\setlength{\tabcolsep}{6pt}
\renewcommand{\arraystretch}{1.15}
\begin{tabular}{@{}lccc@{}}
\toprule
Model & ECtHR-A & SCOTUS & CaseHOLD \\
\midrule
BERT (fine-tuned) & 63.6 & 58.3 & 70.8 \\
RoBERTa (fine-tuned) & 59.0 & 62.0 & 71.4 \\
Legal-BERT (fine-tuned) & 64.0 & 66.5 & 75.3 \\
CaseLaw-BERT (fine-tuned) & 62.9 & 65.9 & 75.4 \\
\midrule
Qwen3-8B (zero-shot) & 88.9 & 83.3 & 83.3 \\
Qwen3-14B (zero-shot) & 82.8 & 73.3 & 76.9 \\
Llama-3.1-8B (zero-shot) & 69.0 & 92.9 & 73.3 \\
Llama-3-8B (zero-shot) & 48.1 & 73.3 & 76.7 \\
Gemma-3-12B (zero-shot) & 66.7 & 83.3 & 73.3 \\
Llama-3.3-70B (zero-shot) & 80.0 & 96.7 & -- \\
Legal-$\Delta$-14B (zero-shot) & 83.3 & 90.0 & 76.7 \\
\bottomrule
\end{tabular}%
\end{table}
\FloatBarrier

\section{CaseHOLD: Full Per-Model Results}
\label{app:casehold-full}

\begin{table*}[h]
\centering
\caption{CaseHOLD, $n=30$ per model, seed$=42$. Answer-swap sensitivity is computed
over the 24 (of 30) cases per model with an automatically detected precedent name.
$^\dagger$Qwen3-8B: all 1{,}000 bootstrap resamples contained zero swap events (18 of 18
usable cases unchanged), so the interval is exactly $[0.0, 0.0]$, not a display error.}
\label{tab:casehold}
\vspace{0.6em}
\footnotesize
\setlength{\tabcolsep}{4pt}
\renewcommand{\arraystretch}{1.15}
\begin{tabular}{@{}lcccccc@{}}
\toprule
Metric & Qwen3-8B & Qwen3-14B & Llama-3.1-8B & Llama-3-8B & Gemma-3-12B & Legal-$\Delta$-14B \\
\midrule
Convergence & 80.0 & 86.7 & 100 & 100 & 100 & 100 \\
Accuracy (of answered cases) & 83.3 & 76.9 & 73.3 & 76.7 & 73.3 & 76.7 \\
Answer-swap sensitivity & 0.0 & 5.6 & 16.7 & 21.7 & 8.3 & 8.3 \\
\quad 95\% bootstrap CI & [0.0, 0.0]$^\dagger$ & [0.0, 16.7] & [4.2, 33.3] & [8.7, 39.1] & [0.0, 20.8] & [0.0, 20.8] \\
\bottomrule
\end{tabular}%
\end{table*}
\FloatBarrier

\section{SCOTUS: Full Per-Model Results}
\label{app:scotus-full}

\begin{table*}[h]
\centering
\caption{SCOTUS Pair A (target area Criminal Procedure, control area Economic
Activity), $n=30$ per model, seed$=42$, all rows in percentages except the bracketed
95\% bootstrap CI (1{,}000 resamples) on verdict-swap sensitivity.}
\label{tab:scotus}
\vspace{0.6em}
\footnotesize
\setlength{\tabcolsep}{4pt}
\renewcommand{\arraystretch}{1.15}
\begin{tabular}{@{}lcccccc@{}}
\toprule
Metric & Qwen3-8B & Qwen3-14B & Llama-3.1-8B & Llama-3-8B & Gemma-3-12B & Legal-$\Delta$-14B \\
\midrule
Convergence (target/control) & 100/100 & 100/100 & 93.3/100 & 100/100 & 100/100 & 100/100 \\
Accuracy (of answered cases) & 83.3 & 73.3 & 92.9 & 73.3 & 83.3 & 90.0 \\
Verdict-swap sensitivity & 43.3 & 30.0 & 53.6 & 33.3 & 60.0 & 70.0 \\
\quad 95\% bootstrap CI & [23.3, 60.0] & [13.3, 46.7] & [35.7, 71.4] & [16.7, 50.0] & [40.0, 76.7] & [53.3, 86.7] \\
Surface mention rate & 100 & 100 & 100 & 66.7 & 96.7 & 93.3 \\
Post-commitment re-engagement & 63.3 & 100 & 23.3 & 16.7 & 50.0 & 0.0 \\
\bottomrule
\end{tabular}%
\end{table*}
\FloatBarrier

\section{ContractNLI: Full Per-Model Results}
\label{app:contractnli-full}

\begin{table*}[h]
\centering
\caption{ContractNLI, target clause \emph{notice on compelled disclosure} against
control clause \emph{no solicitation}, $n=30$ per model, seed$=42$, all rows in
percentages except the \chg{bracketed} 95\% bootstrap CI (1{,}000 resamples) on
verdict-swap sensitivity.}
\label{tab:contractnli}
\vspace{0.6em}
\footnotesize
\setlength{\tabcolsep}{4pt}
\renewcommand{\arraystretch}{1.15}
\begin{tabular}{@{}lcccccc@{}}
\toprule
Metric & Qwen3-8B & Qwen3-14B & Llama-3.1-8B & Llama-3-8B & Gemma-3-12B & Legal-$\Delta$-14B \\
\midrule
Convergence (target/control) & 100/100 & 100/96.7 & 86.7/96.7 & 100/93.3 & 100/100 & 100/100 \\
Accuracy (of answered cases) & 76.7 & 73.3 & 57.7 & 73.3 & 76.7 & 76.7 \\
Verdict-swap sensitivity & 43.3 & 48.3 & 48.0 & 50.0 & 43.3 & 43.3 \\
\quad 95\% bootstrap CI & [26.7, 60.0] & [31.0, 65.5] & [28.0, 68.0] & [32.1, 67.9] & [26.7, 63.3] & [26.7, 60.0] \\
Surface engagement (target) & 100 & 100 & 100 & 93.3 & 100 & 70.0 \\
Post-commitment re-engagement & 60.0 & 83.3 & 66.7 & 60.0 & 80.0 & 3.3 \\
\bottomrule
\end{tabular}%
\end{table*}
\FloatBarrier

\newpage
\section{Ordering of Robustness Metrics}
\label{app:ordering}

\begin{figure}[h]
\centering
\includegraphics[width=0.7\textwidth]{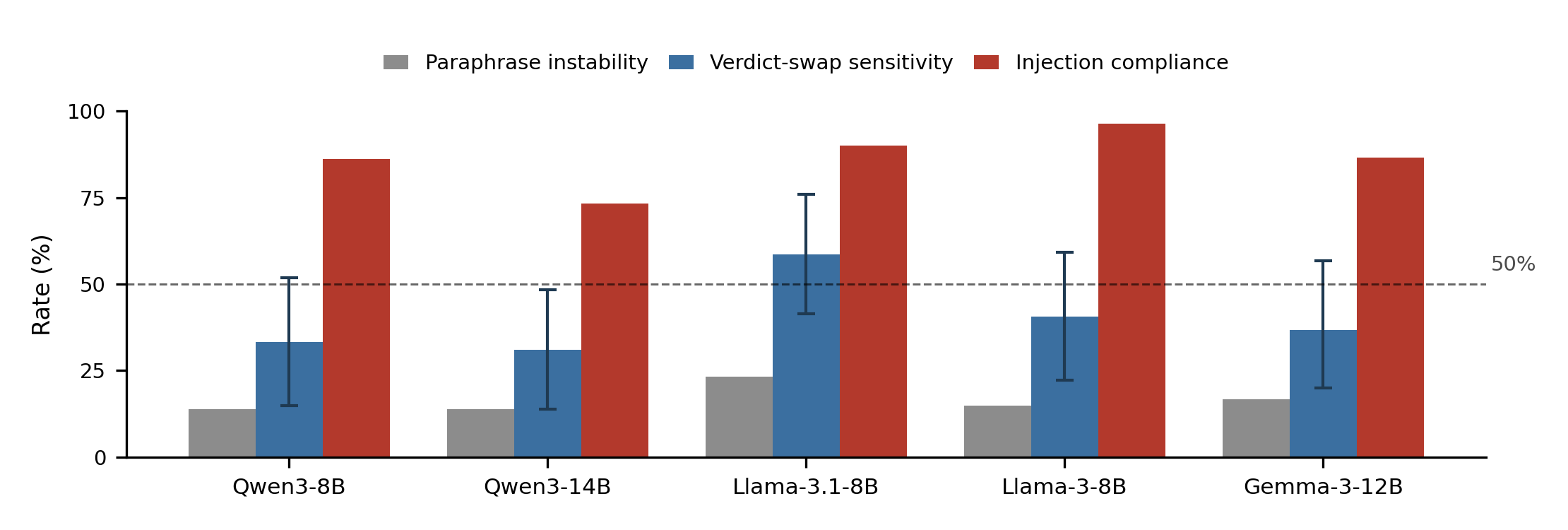}
\caption{ECHR Pair A, $n=30$: paraphrase instability $<$ verdict-swap
sensitivity (95\% CI shown) $<$ injection compliance, without exception, for every
model. A within-model comparison, independent of any cross-model ranking
(Section~\ref{sec:robustness-results}).}
\label{fig:ordering}
\end{figure}
\FloatBarrier

\section{Additional Ablations}
\label{app:ablations}

\paragraph{Is the word-list imbalance a stable property, or merely noise?} The
calibration in Section~\ref{sec:method} corrects for an empirical base rate, which must
itself be trustworthy for the correction to be worth applying. Two independent Qwen3-8B
runs on the identical task, different seeds, some days apart, gave a baseline
\emph{no-violation} share of 79.97\% and 82.11\% (within two points). Across our two
different authority pairs ($n=30$, seed$=42$), the same model's baseline instead moves
seven points, 78.3\% on Pair A against 85.4\% on Pair B, consistent with the imbalance
reflecting genuine per-topic vocabulary rather than run noise, and with our choice to
calibrate separately per run rather than reuse one fixed baseline.

\paragraph{Calibration enabled against disabled: a same-generation paired comparison.}
Using the raw per-sentence distributions saved via \texttt{--save-raw-distributions},
we rescore the identical ECHR Pair A, Qwen3-8B generations ($n=30$) both with and
without calibration%
%\del{, avoiding the sampling confound present in an earlier draft of this ablation}
\chg{, so sampling variation cannot drive any difference}. Aggregate post-commitment re-engagement is unchanged, 53.3\% either way; the exact
committed-at sentence nonetheless differs in 8 of 30 cases (27\%). Calibration thus
shifts individual-case commitment timing without moving the aggregate rate in
Table~\ref{tab:cross-model}, a distinction worth keeping in mind for case-level use of
this pipeline.

\paragraph{Prompting paradigm: native thinking-mode against prompted \texttt{<think>}.}
We re-run Qwen3-8B on identical ECHR Pair A cases with its native thinking-mode toggle
disabled, using the same prompted-\texttt{<think>} instruction given to the other four
models. Surface engagement is unchanged (100\% both ways); convergence rises slightly
(90.0\%$\to$96.7\%), accuracy falls (88.9\%$\to$82.8\%), and post-commitment re-engagement
rises (53.3\%$\to$63.3\%). Committed-before-first-naming stays 0.0\%. Elicitation style
shifts individual measures by ten points or less and does not close the
surface-versus-post-commitment gap central to this paper.

\section{Authority-Pair Replication: Full Per-Model Results}
\label{app:pairrep}

\begin{table}[h]
\centering
\caption{Verdict-swap sensitivity (\%) under two structurally independent
target/control authority pairs per dataset, $n=30$ per cell, seed$=42$. ECHR: Pair A is
Article 8 v. Article 3, Pair B is Article 6 v. Article 14. SCOTUS: Pair A is Criminal
Procedure v. Economic Activity, Pair B is First Amendment v. Civil Rights.}
\label{tab:pair-replication}
\vspace{0.6em}
\footnotesize
\setlength{\tabcolsep}{6pt}
\renewcommand{\arraystretch}{1.15}
\begin{tabular}{@{}lcc@{}}
\toprule
& ECHR & SCOTUS \\
Model & Pair A / Pair B & Pair A / Pair B \\
\midrule
Qwen3-8B & 33.3 / 55.6 & 43.3 / 56.7 \\
Qwen3-14B & 31.0 / 46.4 & 30.0 / 53.3 \\
Llama-3.1-8B & 58.6 / 65.5 & 53.6 / 43.3 \\
Llama-3-8B & 40.7 / 56.7 & 33.3 / 33.3 \\
Gemma-3-12B & 36.7 / 63.3 & 60.0 / 50.0 \\
\bottomrule
\end{tabular}%
\end{table}
\FloatBarrier

\section{Confound-Free Resampling: Full Per-Model Results}
\label{app:differ}

\begin{table}[h]
\centering
\caption{Verdict-swap sensitivity (\%), original sampling against the confound-free
subset (Section~\ref{sec:differ-method}), ECHR and SCOTUS Pair A, $n=30$ per cell,
seed$=42$.}
\label{tab:differ-full}
\vspace{0.6em}
\footnotesize
\setlength{\tabcolsep}{6pt}
\renewcommand{\arraystretch}{1.15}
\begin{tabular}{@{}lcccc@{}}
\toprule
& \multicolumn{2}{c}{ECHR} & \multicolumn{2}{c}{SCOTUS} \\
Model & Orig. & Differ & Orig. & Differ \\
\midrule
Qwen3-8B & 33.3 & 48.1 & 43.3 & 33.3 \\
Qwen3-14B & 31.0 & 51.9 & 30.0 & 36.7 \\
Llama-3.1-8B & 58.6 & 53.3 & 53.6 & 51.7 \\
Llama-3-8B & 40.7 & 21.4 & 33.3 & 46.7 \\
Gemma-3-12B & 36.7 & 43.3 & 60.0 & 66.7 \\
\bottomrule
\end{tabular}%
\end{table}
\FloatBarrier

\section{ContractNLI Confound-Free Resampling: Full Per-Model Results}
\label{app:contractnli-differ}
Collapsing ContractNLI's original
Entailment/Contradiction/NotMentioned labels into a binary YES/NO
(Section~\ref{sec:contractnli-data}) creates a dual-negative risk analogous to
ECHR/SCOTUS: a model answering NO under both the target and control clause is not
necessarily unfaithful if neither clause is genuinely entailed. We address this by
re-deriving each sampled document's control-clause gold label from the source
ContractNLI release under the identical seed (42) used throughout this paper, which
reproduces the exact thirty-document sample per model, and restricting to the twelve
documents where the target and control gold labels genuinely differ; per-model usable
$n$ falls slightly below 12 where one condition did not converge to a scoreable
verdict. Unlike ECHR and SCOTUS, this check is unambiguous: sensitivity rises for every
model on the confound-free subset.

\begin{table}[h]
\centering
\caption{ContractNLI verdict-swap sensitivity (\%), original sampling ($n=30$) against
the confound-free subset (usable $n$ shown per model), seed$=42$. Every model's
sensitivity rises on the confound-free subset.}
\label{tab:contractnli-differ}
\vspace{0.6em}
\footnotesize
\setlength{\tabcolsep}{6pt}
\renewcommand{\arraystretch}{1.15}
\begin{tabular}{@{}lcc@{}}
\toprule
& Original ($n=30$) & Confound-free ($n$) \\
\midrule
Qwen3-8B & 43.3 & 66.7 (12) \\
Qwen3-14B & 48.3 & 66.7 (12) \\
Llama-3.1-8B & 48.0 & 50.0 (10) \\
Llama-3-8B & 50.0 & 63.6 (11) \\
Gemma-3-12B & 43.3 & 66.7 (12) \\
Legal-$\Delta$-14B & 43.3 & 66.7 (12) \\
\bottomrule
\end{tabular}%
\end{table}
\FloatBarrier

\section{Null-Authority Condition: Full Per-Model Results}
\label{app:null-authority}
Compare against real-authority sensitivity
in Table~\ref{tab:cross-model} (ECHR) and Table~\ref{tab:scotus} (SCOTUS,
Appendix~\ref{app:scotus-full}).

The confound cannot be fully closed by design, but three checks narrow it. CaseHOLD
changes only the named precedent while holding the question and all five candidate
holdings fixed, immune to this confound by construction, and its 0.0\% to 21.7\%
sensitivity is, if anything, lower than ECHR/SCOTUS, arguing against a single
nuisance-factor explanation. This null-authority baseline is non-zero on every cell
below, and real-authority sensitivity exceeds it on SCOTUS for four of five models but
is genuinely mixed on ECHR, so we call the confound unresolved rather than closed. The
confound-free subset (Section~\ref{sec:differ-results}) and the paraphrase baseline
(Section~\ref{sec:robustness-results}) point the same way, indicating swap-sensitivity
substantially reflects genuine authority-dependence rather than wording noise or
dual-negative agreement.

\begin{table}[h]
\centering
\caption{Verdict-swap sensitivity (\%) against a null-authority placeholder, $n=30$ per
cell, seed$=42$ (ranges and comparison to real-authority sensitivity in
Section~\ref{sec:discussion}); the SCOTUS exception is Llama-3-8B, the ECHR exceptions
are Qwen3-8B and Qwen3-14B.}
\label{tab:null-authority}
\vspace{0.6em}
\footnotesize
\setlength{\tabcolsep}{6pt}
\renewcommand{\arraystretch}{1.15}
\begin{tabular}{@{}lcc@{}}
\toprule
Model & ECHR (null-authority) & SCOTUS (null-authority) \\
\midrule
Qwen3-8B & 43.3 & 33.3 \\
Qwen3-14B & 51.7 & 26.7 \\
Llama-3.1-8B & 46.4 & 36.7 \\
Llama-3-8B & 17.9 & 36.7 \\
Gemma-3-12B & 10.0 & 43.3 \\
\bottomrule
\end{tabular}%
\end{table}
\FloatBarrier
\section{Pairwise Significance of Model-Ranking Differences}
\label{app:mcnemar}
The bootstrap confidence intervals in Tables~\ref{tab:cross-model} and \ref{tab:scotus}
overlap substantially across models, raising the concern that model-ranking claims
(e.g.\ which model is most or least swap-sensitive) are not statistically distinguishable
at $n=30$. Since every model was run on the identical 30 cases in identical order for a
given dataset and pair, we can test this directly and more powerfully than independent
CIs allow: Tables~\ref{tab:mcnemar-echr} and \ref{tab:mcnemar-scotus} report an exact
McNemar's test (a binomial sign test on the discordant pairs, i.e.\ cases where exactly
one of the two models' verdicts flipped under the swap) for every pair of the six models
on ECHR and SCOTUS Pair A. These $p$-values are uncorrected for the 15 comparisons per
dataset and should be read as exploratory, not as a family-wise-controlled test; only 2
of 15 pairs reach nominal $p<0.05$ on ECHR and 5 of 15 on SCOTUS, and none survives a
Bonferroni correction ($\alpha=0.05/15\approx0.0033$) except the SCOTUS
Qwen3-14B/Legal-$\Delta$-14B pair ($p<0.001$). The two comparisons we draw on most
directly in the main text, %
%\del{Section 6.1}
\chg{Section~\ref{sec:cross-model}}'s contrast of the highest- and lowest-swap-sensitivity models on ECHR
(Llama-3.1-8B-Instruct against Qwen3-14B) and the SCOTUS comparison involving
Legal-$\Delta$-14B and Qwen3-14B, are among the nominally significant pairs, though only
the latter survives Bonferroni correction. Most other
pairwise rankings reported in the text (e.g.\ Section~\ref{sec:pair-replication}'s
per-pair orderings) are not statistically distinguishable at this sample size, and we
treat those specific orderings as a secondary, weakly-supported observation rather than
a claim this paper's central finding depends on; the central finding itself, the gap
between surface engagement and verdict-swap sensitivity, holds within every single
model's own bootstrap CI (Tables~\ref{tab:cross-model}--\ref{tab:contractnli}) and does
not depend on the cross-model ranking being significant.

\begin{table}[h]
\centering
\caption{ECHR Pair A: exact McNemar's test on verdict-swap flip/no-flip outcomes,
every model pair, same 30 cases, seed$=42$. $n_{10}$/$n_{01}$ are discordant-pair
counts; bold $p<0.05$.}
\label{tab:mcnemar-echr}
\vspace{0.6em}
\footnotesize
\setlength{\tabcolsep}{6pt}
\renewcommand{\arraystretch}{1.15}
\begin{tabular}{@{}llccc@{}}
\toprule
Model A & Model B & $n_{10}$ & $n_{01}$ & $p$ \\
\midrule
Qwen3-8B & Qwen3-14B & 7 & 6 & 1.000 \\
Qwen3-8B & Llama-3.1-8B & 4 & 11 & 0.118 \\
Qwen3-8B & Llama-3-8B & 5 & 7 & 0.774 \\
Qwen3-8B & Gemma-3-12B & 9 & 9 & 1.000 \\
Qwen3-8B & Legal-$\Delta$-14B & 7 & 6 & 1.000 \\
Qwen3-14B & Llama-3.1-8B & 2 & 10 & \textbf{0.039} \\
Qwen3-14B & Llama-3-8B & 7 & 10 & 0.629 \\
Qwen3-14B & Gemma-3-12B & 4 & 6 & 0.754 \\
Qwen3-14B & Legal-$\Delta$-14B & 2 & 2 & 1.000 \\
Llama-3.1-8B & Llama-3-8B & 11 & 6 & 0.332 \\
Llama-3.1-8B & Gemma-3-12B & 10 & 4 & 0.180 \\
Llama-3.1-8B & Legal-$\Delta$-14B & 10 & 2 & \textbf{0.039} \\
Llama-3-8B & Gemma-3-12B & 8 & 6 & 0.791 \\
Llama-3-8B & Legal-$\Delta$-14B & 9 & 6 & 0.607 \\
Gemma-3-12B & Legal-$\Delta$-14B & 7 & 6 & 1.000 \\
\bottomrule
\end{tabular}%
\end{table}

\FloatBarrier

\begin{table}[h]
\centering
\caption{SCOTUS Pair A: exact McNemar's test on verdict-swap flip/no-flip outcomes,
every model pair, same 30 cases, seed$=42$. Bold $p<0.05$.}
\label{tab:mcnemar-scotus}
\vspace{0.6em}
\footnotesize
\setlength{\tabcolsep}{6pt}
\renewcommand{\arraystretch}{1.15}
\begin{tabular}{@{}llccc@{}}
\toprule
Model A & Model B & $n_{10}$ & $n_{01}$ & $p$ \\
\midrule
Qwen3-8B & Qwen3-14B & 6 & 2 & 0.289 \\
Qwen3-8B & Llama-3.1-8B & 3 & 5 & 0.727 \\
Qwen3-8B & Llama-3-8B & 7 & 4 & 0.549 \\
Qwen3-8B & Gemma-3-12B & 1 & 6 & 0.125 \\
Qwen3-8B & Legal-$\Delta$-14B & 1 & 9 & \textbf{0.021} \\
Qwen3-14B & Llama-3.1-8B & 4 & 10 & 0.180 \\
Qwen3-14B & Llama-3-8B & 5 & 6 & 1.000 \\
Qwen3-14B & Gemma-3-12B & 0 & 9 & \textbf{0.004} \\
Qwen3-14B & Legal-$\Delta$-14B & 0 & 12 & \textbf{\chg{$<$0.001}} \\
Llama-3.1-8B & Llama-3-8B & 11 & 5 & 0.210 \\
Llama-3.1-8B & Gemma-3-12B & 4 & 7 & 0.549 \\
Llama-3.1-8B & Legal-$\Delta$-14B & 4 & 8 & 0.388 \\
Llama-3-8B & Gemma-3-12B & 2 & 10 & \textbf{0.039} \\
Llama-3-8B & Legal-$\Delta$-14B & 2 & 13 & \textbf{0.007} \\
Gemma-3-12B & Legal-$\Delta$-14B & 2 & 5 & 0.453 \\
\bottomrule
\end{tabular}%
\end{table}
\FloatBarrier

\section{Legal-\texorpdfstring{$\Delta$}{Delta}-14B Merge Validity: Base-Model Comparison}
\label{app:legaldelta-validity}
A concern raised about the Legal-$\Delta$-14B reproduction (Section~\ref{sec:legaldelta-rq},
Section~\ref{sec:limitations}) is that applying a LoRA adapter to a base model it was not
trained on could induce arbitrary feature distortion rather than a coherent legal-RL
adaptation, making conclusions drawn from it invalid regardless of caveat. We test this
directly by running the unmerged base, Qwen2.5-14B-Instruct, zero-shot, on the identical
ECHR Pair A cases, $n=30$, seed$=42$. Table~\ref{tab:legaldelta-validity} compares it
against the merged model. The merge does not degrade task accuracy, 73.3\% (base) versus
83.3\% (merged), and convergence and surface engagement are unchanged (100\% and 96.7--100\%
respectively); a genuinely distorted, arbitrary merge would be a more parsimonious
explanation for \emph{degraded} task performance than for improved accuracy alongside a
consistent, directional behavioural shift. That shift is real: verdict-swap sensitivity
rises (26.7\% to 33.3\%) while post-commitment re-engagement falls sharply (56.7\% to 20.0\%),
the same behavioural-over-mechanistic pattern reported for Legal-$\Delta$-14B relative to
the general-purpose models in Section~\ref{sec:legaldelta-rq}, now visible as a
within-model shift from the identical base under the LoRA adapter alone. This is
consistent with the adapter transferring a coherent, if imperfect, legal-RL adaptation
onto the substituted base rather than corrupting it, and is evidence against, though it
does not conclusively rule out, the merge invalidating the comparisons in
Section~\ref{sec:legaldelta-rq}. We still report all Legal-$\Delta$-14B results as a
best-effort reproduction, not the authors' exact checkpoint.

\begin{table*}[pt]
\centering
\caption{ECHR Pair A, $n=30$, seed$=42$: the unmerged public base model against the
LoRA-merged Legal-$\Delta$-14B, on the identical cases.}
\label{tab:legaldelta-validity}
\vspace{0.6em}
\footnotesize
\setlength{\tabcolsep}{6pt}
\renewcommand{\arraystretch}{1.15}
\begin{tabular}{@{}lcc@{}}
\toprule
Metric & Qwen2.5-14B-Instruct & Legal-$\Delta$-14B \\
& (base, unmerged) & (LoRA-merged) \\
\midrule
Convergence (target/control) & 100 / 100 & 100 / 100 \\
Accuracy (of answered cases) & 73.3 & 83.3 \\
Verdict-swap sensitivity & 26.7 & 33.3 \\
Surface engagement (target) & 100 & 96.7 \\
Post-commitment re-engagement & 56.7 & 20.0 \\
\bottomrule
\end{tabular}%
\end{table*}
\FloatBarrier

\section{Full Limitations}
\label{app:full-limitations}
This appendix gives the complete limitations discussion; Section~\ref{sec:limitations}
in the main body summarises the three most load-bearing points (sample size and
statistical power, the authority-swap confound, and the Legal-$\Delta$-14B reproduction
caveat).

\paragraph{Pair coverage.} Two structurally independent pairs per dataset
(Section~\ref{sec:pair-replication}) mitigate but do not eliminate the concern that
results are an artefact of the specific pairs chosen; pairs with greater doctrinal
similarity than either of ours might show different swap-sensitivity.

\paragraph{What our measures do and do not show.} Surface engagement is a coarse string
match on the authority's name: a model could name an authority without substantively
applying it, or apply it using phrasing our matching misses. Post-commitment
re-engagement is a first attempt at this distinction rather than a complete solution;
specifically, it is a strict lower bound on mechanistic re-engagement, since a model
could in principle attend back to the authority's internal representation without
re-emitting its literal name. We do not run an attention-based check (e.g.\
attention-rollout at the decision layer, or activation-patching; Section~\ref{sec:discussion})
in this work, and we therefore treat the two-phase interpretation as a behavioral
hypothesis rather than a mechanistically established account.

\paragraph{Probe standardisation.} The keyword-based ECHR signal is calibrated per run
and remains a heuristic, not the same probing method as the direct logit-lens readout
used for SCOTUS, ContractNLI, and CaseHOLD. A single-token, YES/NO-style marker, the
requirement for a clean logit-lens readout, is not available for ECHR's natural
VIOLATION/NO VIOLATION verdict, which tokenizes to three and four tokens respectively
across every model we tested; standardising the probe would require re-eliciting every
ECHR generation under a YES/NO marker rather than a lighter reanalysis, which we leave to
future work. Separately, we do not systematically vary how instruction tuning and
tokenizer choice interact to shift exactly when pre-verbalisation commitment forms; we
only observe this as a byproduct of comparing five model families. The logit-lens
readout for CaseHOLD, SCOTUS, and ContractNLI also assumes a single well-behaved first
token per candidate answer, an assumption tokenizer quirks could violate on untested
models.

\paragraph{Model selection and prompting.} All models except Legal-$\Delta$-14B are
general-purpose rather than fine-tuned for legal reasoning, and even Legal-$\Delta$-14B
does not show higher authority-dependence than the general-purpose models as a group
(Section~\ref{sec:legaldelta-rq}). Every result in this paper concerns models explicitly
instructed to name an authority (Section~\ref{sec:method}), so near-universal surface
engagement measures what happens once that instruction is given, not spontaneous
citation behaviour; an unprompted condition would be a useful extension. Model selection
was also partly constrained by local infrastructure: GPT-OSS-20B's native MXFP4
quantisation triggered a CUDA error, and Mistral-7B-Instruct-v0.3's checkpoint was
incomplete, so both were excluded rather than debugged. This exclusion reflects
infrastructure, not a judgment on the models themselves.

\paragraph{Seed sensitivity.} We fix one decoding seed (42) throughout, so individual
outcomes are reproducible, but we have not measured how much a different seed would
shift the reported rates. Earlier, differently-seeded Qwen3-8B pilot runs
(Appendix~\ref{app:ablations}) suggest aggregate rates on a fixed task are stable to
within a few points, though this was not re-verified at $n=30$.

\paragraph{Authority-swap confound.} The authority-swap confound, that changing the
authority also changes the question asked, is not fully eliminable with our current
design (Section~\ref{sec:discussion}). Our null-authority condition narrows it but
leaves a genuinely mixed pattern across ECHR and SCOTUS rather than a clean resolution; a
design that more cleanly isolates baseline prompt-perturbation instability from
authority-specific engagement remains a concrete next step.

\paragraph{Dataset- and method-specific scope.} Our fourth-domain result (ContractNLI,
Section~\ref{sec:contractnli-rq}) uses a single target/control clause pair; the same
pair-coverage limitation noted above for ECHR and SCOTUS applies here too, and we did
not run a second authority pair for this dataset. SCOTUS opinions are truncated to their
first 6{,}000 characters, so the model never sees any reasoning that appears later in an
opinion. Our scale result (Section~\ref{sec:scale-rq}) covers a single 70B model on Pair
A of two datasets, not a cross-model scale sweep or a second authority pair at that
scale. Legal-$\Delta$-14B (Section~\ref{sec:legaldelta-rq}) carries a related caveat: its
published weights are LoRA adapters whose exact training base was never released
publicly, so we merged the adapter onto the closest available public base rather than
the authors' actual checkpoint. We verified that the merged model produces coherent,
correctly-formatted legal reasoning before using it, but this remains a best-effort
reproduction, and its results should be read with that substitution in mind. The
confound-free subset (Section~\ref{sec:differ-results}) and the robustness ablation
(Section~\ref{sec:robustness-results}) are each run on Pair A of one or two datasets with
the same five general-purpose models, not the full cross-pair, cross-dataset battery.

\paragraph{Overall scope.} Our conclusions are restricted to behavioral authority
dependence under the stated prompting and evaluation conditions, rather than spontaneous
citation behavior or a mechanistically complete account of how authorities are
represented internally.

\end{document}